%% file: main.tex
\documentclass[conference,compsoc]{IEEEtran}
\IEEEoverridecommandlockouts
\usepackage{cite}
\usepackage{amsmath,amssymb,amsfonts}
\usepackage{graphicx}
\usepackage{textcomp}
\usepackage{xcolor}
\def\BibTeX{{\rm B\kern-.05em{\sc i\kern-.025em b}\kern-.08em
    T\kern-.1667em\lower.7ex\hbox{E}\kern-.125emX}}
    
\usepackage{booktabs}
\usepackage{bbm}
\usepackage{amsfonts}
\usepackage{amsmath}
\usepackage{mathptmx}
\usepackage{mathrsfs}
\usepackage{multirow}
\DeclareMathAlphabet{\mathcal}{OMS}{cmsy}{m}{n}

\usepackage{amssymb}
\usepackage{algorithm}
\usepackage{algpseudocode}

\graphicspath{ {./fig/} }
\newcommand{\eat}[1]{}

\usepackage{xcolor, soul}
\sethlcolor{yellow}



\begin{document}

\title{Probabilistic QoS Metric Forecasting in Delay-Tolerant Networks\\
Using Conditional Diffusion Models on Latent Dynamics}

\makeatletter
\newcommand{\linebreakand}{%
  \end{@IEEEauthorhalign}
  \hfill\mbox{}\par
  \mbox{}\hfill\begin{@IEEEauthorhalign}
}
\makeatother

\author{\IEEEauthorblockN{Enming Zhang}
\IEEEauthorblockA{\textit{School of Computer Science} \\
\textit{Nanjing University of Posts and Telecommunications}\\
Nanjing, China \\
b20060123@njupt.edu.cn}
\and
\IEEEauthorblockN{Zheng Liu$^{*}$\thanks{*Corresponding author}}
\IEEEauthorblockA{\textit{School of Computer Science} \\
\textit{Nanjing University of Posts and Telecommunications}\\
Nanjing, China \\
zliu@njupt.edu.cn}
\linebreakand
\IEEEauthorblockN{Yu Xiang}
\IEEEauthorblockA{\textit{School of Computer Science} \\
\textit{Nanjing University of Posts and Telecommunications}\\
Nanjing, China \\
1221045920@njupt.edu.cn}
\and
\IEEEauthorblockN{Yanwen Qu}
\IEEEauthorblockA{\textit{School of Computer Information and Engineering} \\
\textit{Jiangxi Normal University}\\
Nanchang, China \\
qu\_yw@jxnu.edu.cn}
}

\maketitle

\begin{abstract}
Active QoS metric prediction, commonly employed in the maintenance and operation of DTN, could enhance network performance regarding latency, throughput, energy consumption, and dependability.
Naturally formulated as a multivariate time series forecasting problem, it attracts substantial research efforts.
Traditional mean regression methods for time series forecasting cannot capture the data complexity adequately, resulting in deteriorated performance in operational tasks in DTNs such as routing. 
This paper formulates the prediction of QoS metrics in DTN as a probabilistic forecasting problem on multivariate time series, where one could quantify the uncertainty of forecasts by characterizing the distribution of these samples. The proposed approach hires diffusion models and incorporates the latent temporal dynamics of non-stationary and multi-mode data into them. 
Extensive experiments demonstrate the efficacy of the proposed approach by showing that it outperforms the popular probabilistic time series forecasting methods.
\end{abstract}


\section{Introduction}

Traditional internet does not function well across extremely long distances or under challenging circumstances. Delay-Tolerant Networks (DTN) pave the way for frequent outages, high error rates, and delays that can last hours or even days by guaranteeing dependable and reliable information transmission. 
DTN often uses a diverse network infrastructure based on various communication modes \cite{ito14} and dominates communications in fields such as vehicular communications \cite{4146881}, wildlife tracking/monitoring networks \cite{10.1145/635508.605408}, and rural area communications \cite{6970359}.
Conventionally, DTN may have frequent interruptions due to crucial issues, including the restrictions of wireless radio's range or base station locations, the budget of sustainable batteries, and the influence of noise disturbance or network attacks \cite{voyiatzis12}, which bring rigorous challenges to consistently stable communication.
Constant monitoring of the quality of service (QoS) in DTN is thus critical.

QoS indicates the performance status of network services, whose metrics are guarantees of network transportation \cite{ROY2018121}. 
Monitoring QoS helps to utilize network resources effectively in the diverse traffic for consistent data delivery of DTN. 
For example,  routing protocols could distribute information more efficiently with lower energy consumption according to the network status. 
Furthermore, QoS requirements vary for different types of network traffic. Monitoring QoS help to prioritize these types of traffic to maximize the network throughput and utilization. 
For example, a system can prioritize packet replication requests by calculating the replication probability based on QoS status related to packet expected transmission delay and node encounter possibility.

%主动Qos预测的应用。

Unlike monitoring QoS metrics, which only reveal the current network status, active QoS metrics prediction helps to enhance the network performance in latency, throughput, energy consumption, and dependability \cite{pundir_quality--service_2021,abdellah_machine_2021,s_yellampalli_applications_2021,ma_research_2021}. The system could prepare solutions to potential QoS metrics changes in the future. 
Formulating QoS metrics prediction as a multivariate time series forecasting problem is straightforward, in which the predictions are the future spot values of QoS metrics. The autoregression model is the foundation of most existing methods for time series forecasting. These methods focus on providing accurate predictions by minimizing the deviations between the predicted spot values and the ground truth. The predictions are the mean regression values since the deviations are measured by the mean squared error (MSE) or other similar accuracy measurements.

Mean regression methods work well when there is only one mode in the distribution of the response variable. However, if the data contains multiple modes, they are unable to capture the full complexity of the data because they may result in estimates that are not representative of the data or fail to capture the full range of variation in the response variable \cite{hyndman_athanasopoulos_2021}. 
The results may sometimes be misleading, even though the outputs are the most precise regarding error measurements. In the maintenance and operation tasks of DTN, such as routing, it is necessary to quantify the forecasting uncertainty, while deterministic mean regression methods cannot satisfy this requirement.

% benifits of probabilistic forecasting for DTN
To address the above concerns, we formulate the QoS metrics prediction in DTN as a probabilistic forecasting problem on multivariate time series in this study. Probabilistic forecasting quantifies the variance of the future spot values by identifying their distribution, which could contribute to managing uncertainty and mitigating the impact of delays in DTN. It also helps to allocate resources, adjust schedules, and reroute traffic, improving overall network performance.  
%
% issues of probabilistic forecasting for DTN
The statistical characteristics of the QoS time series may change over time. For example, the underlying patterns of the QoS time series are different when the network is highly congested or not. 
The challenge is developing an accurate and robust model to capture the dynamics of non-stationary and multi-mode QoS time series. 

%\cite{pmlr-v37-sohl-dickstein15}. 
Recently, the success of generative models has verified their ability to produce realistic images. This study explores their potential in probabilistic time series forecasting under the context of DTN, specifically diffusion models \cite{ho_denoising_2020}.
We propose to build a diffusion model on the latent temporal dynamics for probabilistic time series forecasting. By sampling the context-sensitive future values from diffusion models, we can quantify their distribution. 
We incorporate encoded latent temporal dynamics from contextual subsequences to address the above challenge. The latent dynamics help the diffusion model deal with non-stationary characteristics. Furthermore, it could tweak the multiple modes in the distribution toward the more probable pattern.
The contributions of this paper are summarized as follows: 
\begin{itemize}
\item We identify the issues in the existing methods for active QoS metric prediction in DTN and formulate it as a probabilistic time series forecasting problem that allows us to quantify the uncertainty of the predictions. 
\item We extend the popular diffusion model to infer the samples for the prediction distributions. We incorporate the latent contextual dynamics in the diffusion model to improve the model's adaptability to non-stationary and multi-mode time series.
\item We conduct extensive experiments to compare our method with the state-of-the-art approaches and demonstrate its advantages. Considering the contextual dynamics in the diffusion model leads to more accurate and reliable predictions for QoS metric prediction in DTN, as shown in the results.
\end{itemize}

\section{Probabilistic forecasting of QoS metrics}
\label{sec:prob}
\input{prob}

\section{The Framework}
\label{sec:fw}

%Existing probabilistic forecasting methods include conditional quantile regression \cite{} and conditional expectile regression \cite{}, as well as Bayesian-based approaches which employ variational approximations \cite{}, expectation propagation \cite{} or probabilstic machine learning strategies \cite{}. 

\eat{
In the image generation domain, GAN \cite{goodfellow_generative_2014}, and VAE \cite{Kingma2014} have enormous potential for producing realistic images. However, the training process of GAN is unstable, and VAE usually relies on a surrogate loss, resulting in deteriorated quality. Recently, another promising generative model, the denoising diffusion model \cite{ho_denoising_2020}, beats GAN and VAE and generates images so successfully that they are becoming popular in both the academic and the industry.
}

Our proposed probabilistic time series forecasting framework is based on the denoising diffusion model \cite{ho_denoising_2020}, which makes it convenient to qualify the prediction distributions by inferring samples. This section will introduce the overall framework and the individual components.

\begin{figure*}[t]
    \centering
    \includegraphics[width=0.95\textwidth]{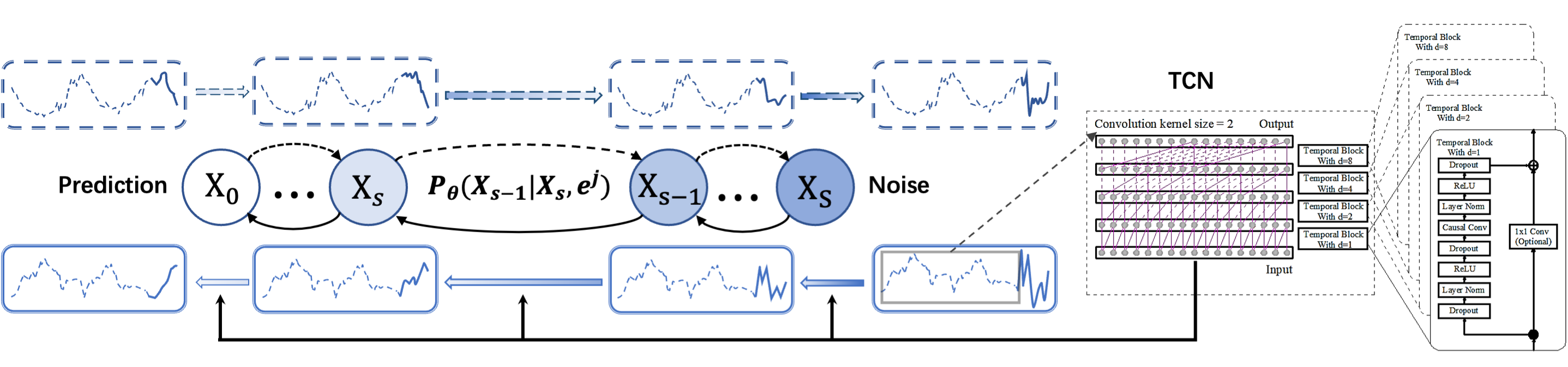}
    \caption{The diffusion (forward) and denoising (reverse) processes of the conditioned diffusion models.}
    \label{fig:all}
\end{figure*}

\subsection{Diffusion models}
\label{sec:diffusion}
We first introduce the diffusion models in the context of time series forecasting for easy understanding and the corresponding diffusion and denoising processes.

Recall $\mathbf{X}=(\mathbf{x}^{1},\mathbf{x}^{2},...,\mathbf{x}^{t})\in \mathbb{R}^{m\times t}$ denote the existing multivariate time series of length $t$ where $ \mathbf{x}^i=(x^{i,1},x^{i,2},\cdots,x^{i,m})$.
Here, we move subscripts to superscripts for convenience in the following explanation of diffusion models. Without loss of generality, the probabilistic time forecasting problem is to sample the future values denoted as $\widetilde{\mathbf{X}}=(\mathbf{x}^{t+1}, \mathbf{x}^{t+2},..., \mathbf{x}^{t+p}) \in \mathbb{R}^{n \times p}$, where $n \leq m$. 
Let $\mathbf{x}^j$ be a data instance from the real data distribution $q(\mathbf{x}^j)$ of the future values at time $j$. In the following equations, $\mathbf{x}^j$ is denoted as $\mathbf{x}$ unless specified for notation simplicity.

There are two processes in diffusion models: the forward diffusion process and the reverse denoising process, as demonstrated in Fig.~\ref{fig:all}.
In the forward diffusion process, a small amount of Gaussian noise is added to the instance $\mathbf{x}$ step by step up to $S$ steps, producing a sequence of noisy instances. The step sizes are controlled by a variance schedule $\{\beta_s \in (0,1)\}_{s=1}^S$.

$q(\mathbf{x}_s \vert \mathbf{x}_{s-1})$ in Eq.~\ref{eq:q} represents the instance distribution at each step.
\begin{equation}
\label{eq:q}
q(\mathbf{x}_s \vert \mathbf{x}_{s-1}) = \mathcal{N}(\mathbf{x}_s; \sqrt{1 - \beta_s} \mathbf{x}_{s-1}, \beta_s\mathbf{I}).
\end{equation}
Let $\mathbf{x}_0$ be the real instance at the very beginning, then the joint distribution under the Markov chain assumption is
\begin{equation}
\label{eq:q2}
q(\mathbf{x}_{1:S} \vert \mathbf{x}_0) = \prod^S_{s=1} q(\mathbf{x}_s \vert \mathbf{x}_{s-1}). \quad
\end{equation}
As $s$ increases, $\mathbf{x}_0$ loses its distinguishing characteristics and approaches a noise distribution.

If $q(\mathbf{x}_{s-1} \vert \mathbf{x}_s)$ is available, it is possible to reverse the above process and generate new samples from a Gaussian noise $\mathcal{N}(\mathbf{x}_S; \mathbf{0},\mathbf{I})$.
The actual $q(\mathbf{x}_{s-1} \vert \mathbf{x}_s)$ is difficult to acquire. However, we could learn a model $p_\theta$ to approximate $q(\mathbf{x}_{s-1} \vert \mathbf{x}_s)$ for the reverse process in Eq.~\ref{eq:rev}.
\begin{equation}
\label{eq:rev}
p_\theta(\mathbf{x}_{0:S}) = p(\mathbf{x}_S) \prod^S_{s=1} p_\theta(\mathbf{x}_{s-1} \vert \mathbf{x}_s).
\end{equation}
When $\beta_s$ is small, we can model the conditional probabilities by Gaussian as follows \cite{ho_denoising_2020}: 
\begin{equation}
\label{eq:p}
p_\theta(\mathbf{x}_{s-1} \vert \mathbf{x}_s) = \mathcal{N}(\mathbf{x}_{s-1}; \boldsymbol{\mu}_\theta(\mathbf{x}_s, s), \sigma_\theta(\mathbf{x}_s, s)\mathbf{I}).
\end{equation}

With the help of variational lower bound \cite{weng2021diffusion}, we have 
\begin{equation}
\begin{aligned}
- \log  p_\theta & (\mathbf{x}_0)  \\
\leq &- \log p_\theta(\mathbf{x}_0) + D_\text{KL}(q(\mathbf{x}_{1:S}\vert\mathbf{x}_0) \| p_\theta(\mathbf{x}_{1:S}\vert\mathbf{x}_0) ) \\
= &\mathbb{E}_q \Big[ \log \frac{q(\mathbf{x}_{1:S}\vert\mathbf{x}_0)}{p_\theta(\mathbf{x}_{0:S})} \Big]. \\
\end{aligned}
\end{equation}
The log-likelihood of $\mathbf{x}_0$ should be as large as possible, so  in order to minimize the negative log-likelihood, we could minimize the following loss function.
\begin{equation}
\label{eq:loss}
L = \mathbb{E}_{q(\mathbf{x}_{0:S)})} \Big[ \log \frac{q(\mathbf{x}_{1:S}\vert\mathbf{x}_0)}{p_\theta(\mathbf{x}_{0:S})} \Big] \geq - \mathbb{E}_{q(\mathbf{x}_0)} \log p_\theta(\mathbf{x}_0).
\end{equation}
Unfortunately, the loss function is not analytically computable, so parametrization is necessary \cite{pmlr-v37-sohl-dickstein15}.
Eq.~\ref{eq:loss} could be rewritten as 
\begin{equation}
\begin{aligned}
L  = &\mathbb{E}_{q(\mathbf{x}_{0:S)})} \Big[ \log \frac{q(\mathbf{x}_{1:S}\vert\mathbf{x}_0)}{p_\theta(\mathbf{x}_{0:S})} \Big] \\
 =& \mathbb{E}_q [D_\text{KL}(q(\mathbf{x}_S \vert \mathbf{x}_0) \parallel p_\theta(\mathbf{x}_S)) \\
 &+ \sum_{t=2}^T D_\text{KL}(q(\mathbf{x}_{s-1} \vert \mathbf{x}_s, \mathbf{x}_0) \parallel p_\theta(\mathbf{x}_{s-1} \vert\mathbf{x}_s)) \\
 &- \log p_\theta(\mathbf{x}_0 \vert \mathbf{x}_1) ],
\end{aligned}
\end{equation}
where $D_\text{KL}(q(\mathbf{x}_S \vert \mathbf{x}_0) \parallel p_\theta(\mathbf{x}_S))$ is constant and thus ignored, and $-\log p_\theta(\mathbf{x}_0 \vert \mathbf{x}_1)$ is handled by a separate discrete decoder.
Now the key is to parameterize $D_\text{KL}(q(\mathbf{x}_{s-1} \vert \mathbf{x}_s, \mathbf{x}_0) \parallel p_\theta(\mathbf{x}_{s-1} \vert\mathbf{x}_s))$.
As in Eq.~\ref{eq:p}, 
\begin{equation}
\boldsymbol{\mu}_\theta(\mathbf{x}_s, s) = \frac{1}{\sqrt{\alpha_s}} \Big( \mathbf{x}_s - \frac{1 - \alpha_s}{\sqrt{1 - \bar{\alpha}_s}} \boldsymbol{\epsilon}_\theta(\mathbf{x}_s, s) \Big),
\label{eq:s}
\end{equation}
so 
\begin{equation}
\begin{aligned}
L_s 
&= \mathbb{E}_{\mathbf{x}_0, \boldsymbol{\epsilon}} \Big[\frac{ (1 - \alpha_s)^2 }{2 \alpha_s (1 - \bar{\alpha}_s) \| \sigma_\theta(\mathbf{x}_s, s)\|^2_2} \\
&\quad \quad \quad \quad \quad \|\boldsymbol{\epsilon}_s - \boldsymbol{\epsilon}_\theta(\sqrt{\bar{\alpha}_s}\mathbf{x}_0 + \sqrt{1 - \bar{\alpha}_s}\boldsymbol{\epsilon}_s, s)\|^2 \Big] .
\end{aligned}
\end{equation}

\eat{
\begin{equation}
\begin{aligned}
L_s 
&= \mathbb{E}_{\mathbf{x}_0, \boldsymbol{\epsilon}} \Big[\frac{1}{2 \| \sigma_\theta(\mathbf{x}_s, s)\mathbf{I}\|^2_2} \| \tilde{\boldsymbol{\mu}}_s(\mathbf{x}_s, \mathbf{x}_0) - {\boldsymbol{\mu}_\theta(\mathbf{x}_s, s)} \|^2 \Big] \\
&= \mathbb{E}_{\mathbf{x}_0, \boldsymbol{\epsilon}} \Big[\frac{ (1 - \alpha_s)^2 }{2 \alpha_s (1 - \bar{\alpha}_s) \| \sigma_\theta(\mathbf{x}_s, s)\|^2_2} \\
&\quad \quad \quad \quad \quad \|\boldsymbol{\epsilon}_s - \boldsymbol{\epsilon}_\theta(\sqrt{\bar{\alpha}_s}\mathbf{x}_0 + \sqrt{1 - \bar{\alpha}_s}\boldsymbol{\epsilon}_s, s)\|^2 \Big] .
\end{aligned}
\end{equation}
}

Let $\sigma_\theta(\mathbf{x}_s, s)$ be $\beta_t$, which is fixed to be a constant. Ho et al. \cite{ho_denoising_2020} also found that a simple version of the loss function by removing the weighting term could obtain better training results. 
\begin{equation}
\begin{aligned}
 L_s& ^\text{simple} = \\ 
& \mathbb{E}_{s \sim [1, S], \mathbf{x}_0, \boldsymbol{\epsilon}_s} \Big[\|\boldsymbol{\epsilon}_s 
-\boldsymbol{\epsilon}_\theta(\sqrt{\bar{\alpha}_s}\mathbf{x}_0 + \sqrt{1 - \bar{\alpha}_s}\boldsymbol{\epsilon}_s, s)\|^2 \Big].
\end{aligned}
\end{equation}
To summarize, the loss function is 
\begin{equation}
\begin{aligned}
L_\text{simple} = L_s^\text{simple} + C,
\end{aligned}
\end{equation}
where $C$ is a constant.

\subsection{Conditional diffusion models on contextual subsequences}
\label{sec:cond}
When applying diffusion models to probabilistic time series forecasting, the forward process remains the same. Despite the ability of diffusion models to address overfitting and generalization concerns due to their high expressiveness, a notable challenge emerges in the reverse process of generating samples from the distribution of future values. This procedure could potentially yield an uncertain distribution that lacks reliability and accuracy, particularly when dealing with complex real-world time series exhibiting substantial variability and uncertainty.
In the forecasting task, more tractable and accurate prediction is expected. A commonly used solution is conditional diffusion models, in which features and supplementary conditional information are jointly fed into the model to provide valuable cues that assist the diffusion model in generating more reliable forecasts.

It is easy to see that the reverse denoising process could generate samples from future value distribution only if $p_\theta(\mathbf{x}_{s-1} \vert \mathbf{x}_s)$ is conditioned on the temporal dynamics of the contextual subsequences, which serve as trustworthy references for the denoising process. This, in turn, enables more precise modeling of the temporal stochasticity inherent in DTN data and improves understanding of the underlying data distribution.
Let $e^j$ denote the temporal dynamic at time $j$, then $p_\theta(\mathbf{x}_{s-1} \vert \mathbf{x}_s)$ conditioned on $e^j$ is denoted as $p_\theta(\mathbf{x}_{s-1} \vert \mathbf{x}_s, e^j)$. The corresponding conditional version of Eq.~\ref{eq:rev} is 
\begin{equation}
\label{eq:rev2}
p_\theta(\mathbf{x}_{0:S}) = p(\mathbf{x}_S) \prod^S_{s=1} p_\theta(\mathbf{x}_{s-1} \vert \mathbf{x}_s, e^j).
\end{equation}
When $\beta_s$ is small, we can model the conditional probabilities by Gaussian as follows: 
\begin{equation}
\label{eq:p2}
p_\theta(\mathbf{x}_{s-1} \vert \mathbf{x}_s, e^j) = \mathcal{N}(\mathbf{x}_{s-1}; \boldsymbol{\mu}_\theta(\mathbf{x}_s, s, e^j), \sigma_\theta(\mathbf{x}_s, s, e^j)\mathbf{I}).
\end{equation}
The corresponding loss function is 
\begin{equation}
\label{eq:finalloss}
\begin{aligned}
L_s&^\text{simple} = \\ 
&\mathbb{E}_{s \sim [1, S], \mathbf{x}_0, \boldsymbol{\epsilon}_s} \Big[\|\boldsymbol{\epsilon}_s - 
\boldsymbol{\epsilon}_\theta(\sqrt{\bar{\alpha}_s}\mathbf{x}_0 + \sqrt{1 - \bar{\alpha}_s}\boldsymbol{\epsilon}_s, s), e^j\|^2 \Big].
\end{aligned}
\end{equation}

\subsection{Latent temporal dynamics}
\label{sec:dynamics}
\input{tcn}

%TCN

\subsection{Training and sampling from the model}
\label{sec:archtecture}
%include model archtecture.
The principal component of the architecture is the use of transformers instead of dilated convolutions layers as an ideal main computational building block of the model, which is known for the ability to attend to specific parts of the input sequence rather than considering the entire sequence equally.
The transformer plays a crucial role in capturing the long-term dependencies within the time series. It enables the dynamic weighting of feature values at different time steps and takes into account the diffusion embeddings. The attention mechanism employed in the transformer is employed to capture the inter-dependencies between different variables at each time step \cite{zhou2021informer}. The incorporation of a residual structure further enhances the model capacity by facilitating gradient propagation and enabling better prediction performance.

\begin{figure}[t]
    \centering
    \includegraphics[width=0.8\columnwidth]{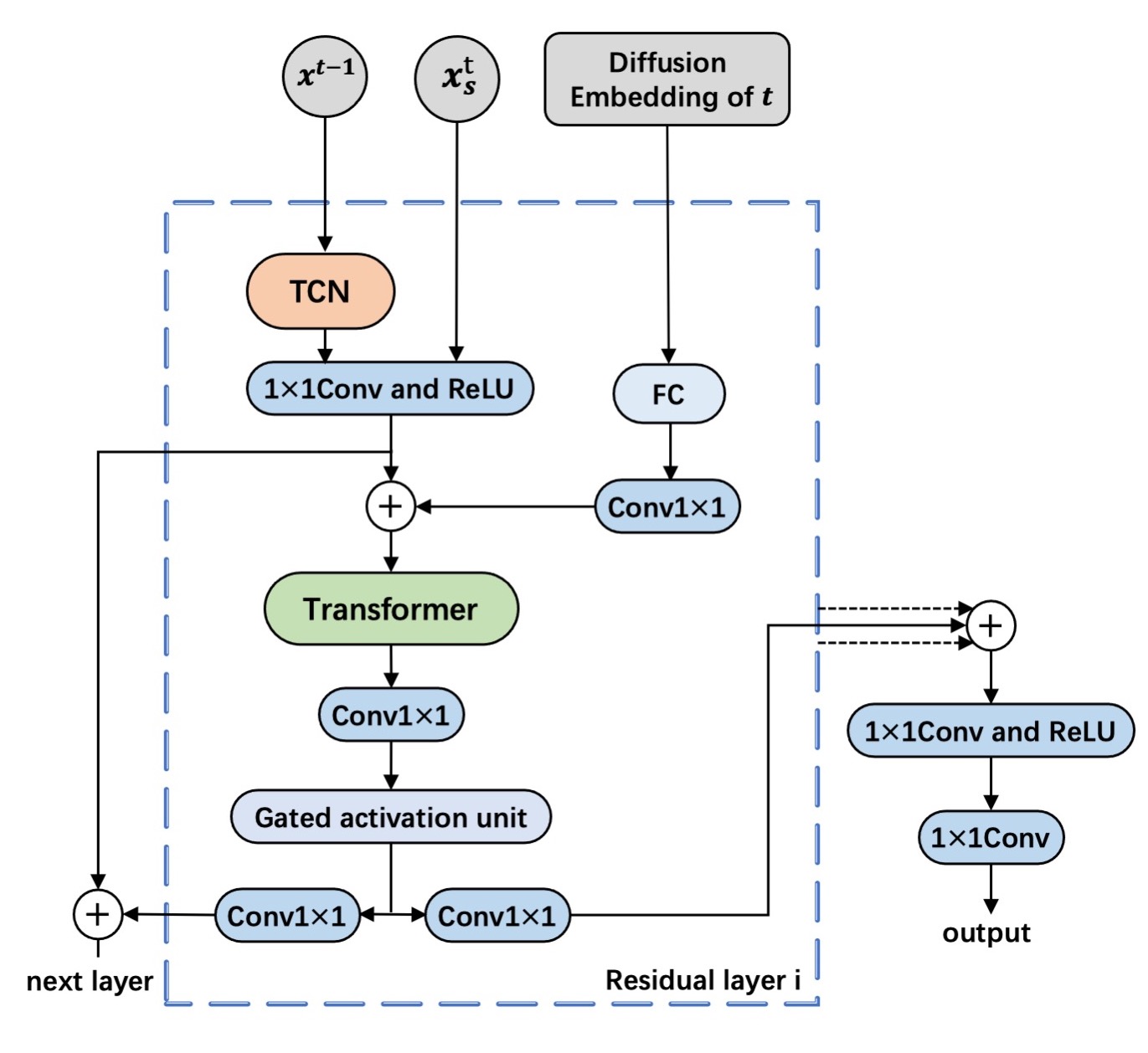}
    \caption{The overall architecture}
    \label{fig:arch}
\end{figure}

\eat{
\begin{algorithm}[t]
\caption{Training procedure}
\label{algo:t}
    \begin{algorithmic}[1]
        \State Sample $\mathbf{x}_0^j$ from training data, $\epsilon_s$ from $\mathcal{N}(0,I)$, and $s$ from $\text{Uniform}({1..S})$,
        \Repeat
        \State Do gradient decedent on  
        \Statex \quad \quad \quad \quad$\nabla\|\boldsymbol{\epsilon}_s - \boldsymbol{\epsilon}_\theta(\sqrt{\bar{\alpha}_t}\mathbf{x}_0 + \sqrt{1 - \bar{\alpha}_s}\boldsymbol{\epsilon}_s, s), e^j\|^2 $\,
        \Until A fixed number of iterations or converged
    \end{algorithmic}
\end{algorithm}

\begin{algorithm}[t]
\caption{Sampling predictions}
\label{alg:s}
    \begin{algorithmic}[1]
        \State Sample $\mathbf{x}_S^j$ from noise distribution $\mathcal{N}(0,I)$
        \For {$s=S$ \textbf{to} 1}  
        \State $\mathbf{x}_{s-1}^j=\frac{1}{\sqrt{\alpha_s}}\left(\mathbf{x}_s^j-\frac{\beta_s}{\sqrt{1-\bar{\alpha}_s}} \epsilon_\theta\left(\mathbf{x}_s^j, s, e^j\right)\right)+\sqrt{\Sigma_\theta} \mathbf{z}$
    \end{algorithmic}
\end{algorithm}
}

The overall architecture of the proposed framework is presented in Fig.~\ref{fig:arch}, which shows a single residual block and the output obtained from the sum of all skip-connections \cite{kong2020diffwave}.
Training is performed by randomly sampling context and adjoining prediction-sized windows from the training time series data and minimizing the loss function in Eq.~\ref{eq:finalloss} by optimizing the parameters of ${\epsilon}_\theta$ and the TCN model.
With the trained conditional diffusion model, one can obtain a sample of the prediction for the next few steps by sampling an initial noise and denoising based on Eq.~\ref{eq:s}.

\eat{
The training procedures are presented in Algorithm \ref{algo:t}.
}

\section{Experimental Evaluation}
\label{sec:exp}
 
\subsection{Datasets and metrics}
We assess the viability of our proposed approach using a publicly available sensor dataset, encompassing metrics such as internet latency alongside various other pertinent features. This dataset\footnote{https://www.kaggle.com/datasets/johntrunix/home-sensordata} was meticulously assembled by capturing feature data, including internet latency and other features, at one-minute intervals spanning over a duration of more than 100 days. As an example, we present a line graph in Fig.~\ref{fig:dataset} that examines internet latency patterns on October 6, 2021. The x-axis represents time in hours (0-23), while the y-axis shows latency in milliseconds. The graph displays varying latency levels, with peaks up to 50ms and troughs near 17ms. As the graph clearly demonstrates, latency exhibits strong fluctuations, which validates our reasoning and the necessity for formulating it as a probabilistic forecasting problem rather than using a mean regression method.

To facilitate our experimentation, we partitioned the dataset into two distinct segments, denoted as D1 and D2. This partitioning was guided by specific time ranges and underlying distributions inherent in the data.
The composition of D1 comprises approximately 80,000 sequential timestamps, whereas D2, distinct in its value ranges and distributions, encompasses around 40,000 temporal instances. Our method revolves around leveraging historical subsequences of all features, encompassing the most recent 120 timestamps, to make predictions about the forthcoming latency values within the subsequent ten timestamps.
The dataset's comprehensive nature, containing internet latency data as well as complementary features, mirrors the real-world complexity of delay-tolerant networks. The division of the dataset into D1 and D2 further enables us to demonstrate the method's robustness across distinct data distributions and ranges. By relying on historical feature subsequences to predict latency, we showcase the versatility of our model in tackling real-time prediction scenarios.

\eat{
We divide the data set into two separate parts, i.e., D1 and D2, based on value ranges and distributions. The length of D1 is around 80,000 time steps, while the length of D2 is around 40,000 time steps.} 

\begin{figure}[t]
    \centering
    \includegraphics[width=0.8\columnwidth]{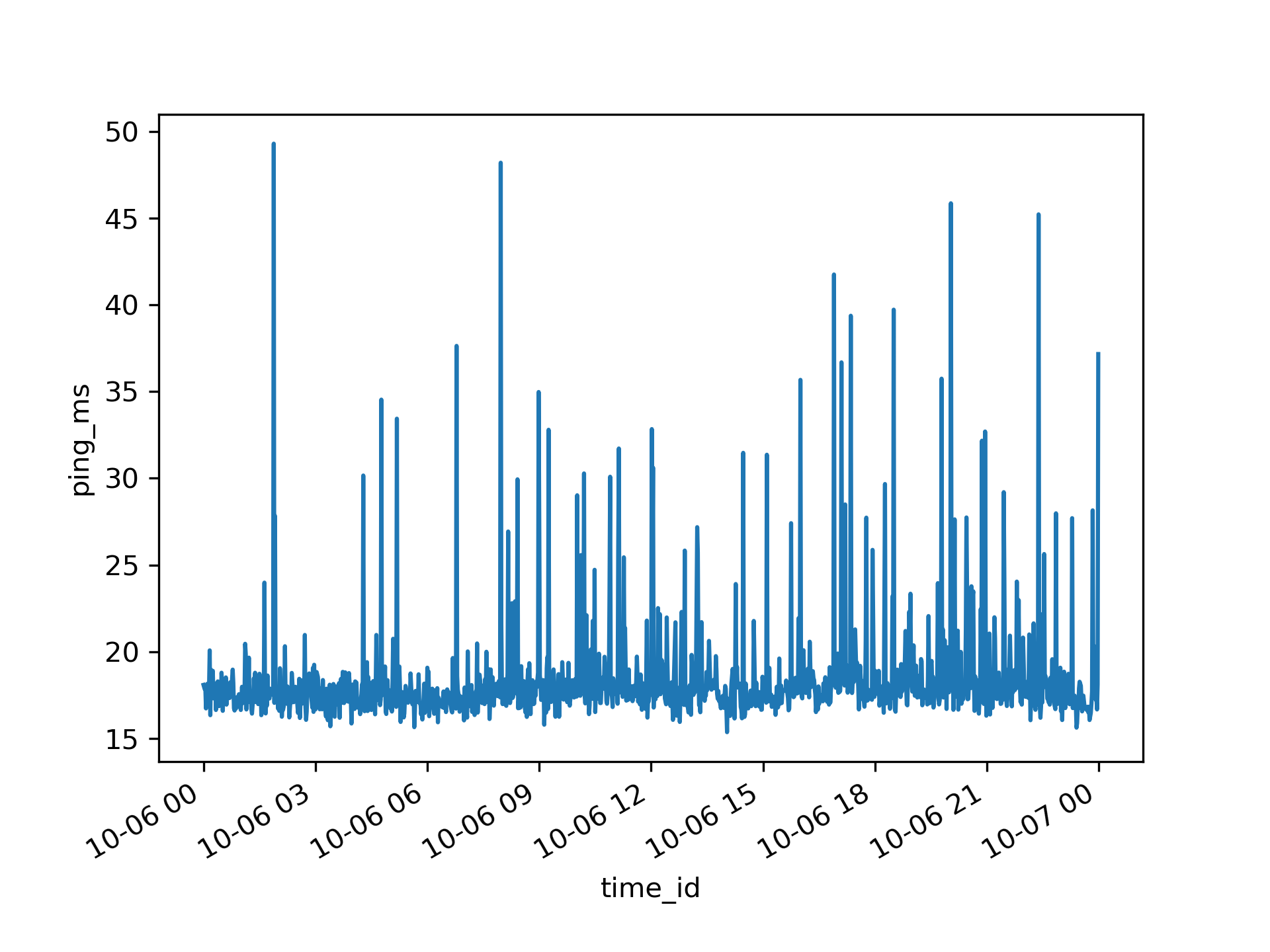}
    \caption{Hourly internet latency variations on October 6, 2021.}
    \label{fig:dataset}
\end{figure}

We report the experimental results using two commonly used scale-dependent measures, i.e., mean absolute errors (MAE) and mean squared errors (MSE). Here the forecasting values are the average of all inferred samples. 
We use the Continuous Ranked Probability Score (CRPS) to assess the quality of probabilistic forecasts, which generalizes MAE and measures the compatibility of a cumulative distribution function $F$ with an observation $x$:
\begin{equation}
    \operatorname{CRPS}(F, x)=\int_{-\infty}^{\infty}(F(y)-\mathbbm{1}(x \leq y))^2 d y
\end{equation}
where $\mathbbm{1}$ is the indicator function which is one if $y \leq x$ and zero otherwise. $F(y)=P(x \leq y)$ is approximated by the 100 inferred samples. A smaller CRPS means that the distribution of forecasts is closer to the true distribution \cite{hersbach2000decomposition}.

\eat{
Aside from CRPS, the two most commonly used scale-dependent measures, absolute errors (MAE) and squared errors (MSE), are also used to assess the prediction results.  Considering we have a random sample of size $n$, the average of them is regarded as the estimated values, so they can be computed as:
\begin{equation}
\text { MSE(Y) }=E\left[\left(\frac{1}{n} \sum_{i=1}^{n} Y_{i}-\hat{Y}\right)^{2}\right]
\end{equation}
\begin{equation} 
\text { MAE(Y) }=E\left[\left|\frac{1}{n} \sum_{i=1}^{n} Y_{i}-\hat{Y}\right|\right]
\end{equation}
where $\hat{Y}$ is the true values, and $Y_{i}$ is the $i_{th}$ sample values.
}

We compare the proposed framework, DiffTCN, with four popular probabilistic time series forecasting methods as the baseline. 
\begin{itemize}
  \item DeepAR\cite{SALINAS20201181}: DeepAR is a deep auto-regression methodology for producing accurate probabilistic forecasts using recurrent network models. It learns a global model from related time series, handles widely-varying scales through rescaling and velocity-based sampling, generates calibrated probabilistic forecasts with high accuracy, and is able to learn complex patterns such as seasonality and uncertainty growth over time from the data.  
  
  \item DeepFactor\cite{pmlr-v97-wang19k}: Deep Factors is a hybrid model that combines the strengths of classical time series models and deep neural networks to produce accurate and scalable probabilistic forecasts. It assumes that all the time series are driven by a small number of dynamic (latent) factors and uses a global-local structure to extract complex non-linear patterns globally while capturing individual random effects for each time series locally. 
  
  \item TimeGrad\cite{rasul2021autoregressive}: The Autoregressive Denoising Diffusion Models proposes a new state-of-the-art approach for multivariate probabilistic forecasting using diffusion probabilistic models and Langevin sampling. The method models the full predictive distribution and incorporates inductive bias in the form of multivariate probabilistic methods to provide accurate forecasts.
  
  \item Diffusion\cite{ho_denoising_2020}: the original vanilla diffusion model described in Section \ref{sec:diffusion}.
\end{itemize}

\subsection{Implementation details}
Adam optimizer was employed with an initial learning rate of 0.001 during training. The number of diffusion steps is 50. The minimum noise level is $\beta_1 = 0.0001$, and the maximum noise level is $\beta_T = 0.5$. The batch size is 64 by selecting random subsequences (with possible overlaps) from all the possible time positions. The number of inferred samples is 100 to approximate the distributions.
The dilation of the TCN layer with a bidirectional dilated convolution is 2.
To encode the diffusion step $s$, we employ the position embeddings from the Transformer\cite{NIPS2017_3f5ee243}, with the ${d}_{model}=128$.
Regarding Transformer layers, we created a 1-layer TransformerEncoder in PyTorch, which consists of a multi-head attention layer, fully-connected layers, and layer normalization.
All experiments were conducted on a PC with a single Nvidia 3090 GPU (24GB memory). The hyper-parameters and detection thresholds of the baseline methods are set based on the information provided in their respective original papers.

\eat{
Adam optimizer was utilized with an initial learning rate of 0.001 during the training process. Additionally, we set the number of diffusion steps to 50. The batch size is 64 by selecting random windows (with possible overlaps) from all the possible time positions in the time series dataset, with the context size set to the number of prediction horizons. For testing, we employ a rolling window for prediction, beginning with the previous context window history before the commencement of the prediction. The number of samples is 100 to obtain the approximate distributions.

To encode the diffusion step $s$, we employ the position embeddings of the Transformer, with the ${d}_{model}=128$.
The dilation of the TCN layer with a bidirectional dilated convolution is set to 2. Regarding Transformer layers, we created a 1-layer TransformerEncoder in PyTorch, which consists of a multi-head attention layer, fully-connected layers, and layer normalization.
All experiments run on a single Nvidia 3090 GPU with 24GB of memory.
}
\eat{
I added a header row for distinguishing the two datasets and included both MSE and MAE in the combined table. The label has also been updated to "table:combined" for referencing purposes. Feel free to adjust the caption and label according to your needs.}

\subsection{Experimental results}
%The reported values are the average performance of ten trials.
\begin{table*}[t]
\centering
\caption{The MAE and MSE results of DiffTCN and baseline methods on Datasets D1 and D2 at four forecast horizons (1, 4, 7, and 10). (The reported results are the average performance of ten trials.)}
\begin{tabular}{ccccc|ccc}
\toprule
\multirow{2}{*}{\textbf{Dataset}} & \multirow{2}{*}{\textbf{Horizon}} & \multicolumn{3}{c}{\textbf{MSE}} & \multicolumn{3}{c}{\textbf{MAE}} \\ \cmidrule(lr){3-8}
                 &                  & \textbf{Diffusion} & \textbf{TimeGrad} & \textbf{DiffTCN(Our Method)} & \textbf{Diffusion} & \textbf{TimeGrad} & \textbf{DiffTCN(Our Method)} \\ \midrule
\multirow{4}{*}{\textbf{D1}} & \textbf{1}  & 2.750($\pm 0.312$) & 2.372($\pm 0.084$) & \textbf{1.959($\pm 0.033$)} & 1.150($\pm 0.021$) & 1.020($\pm 0.011$) & \textbf{0.749($\pm 0.014$)} \\
                 & \textbf{4}  & 2.556($\pm 0.192$) & 2.390($\pm 0.118$) & \textbf{1.938($\pm 0.014$)} & 1.012($\pm 0.019$) & 0.884($\pm 0.012$) & \textbf{0.756($\pm 0.008$)} \\
                 & \textbf{7}  & 2.823($\pm 0.230$) & 2.407($\pm 0.241$) & \textbf{1.914($\pm 0.029$)} & 1.170($\pm 0.015$) & 1.047($\pm 0.020$) & \textbf{0.752($\pm 0.007$)} \\
                 & \textbf{10} & 2.594($\pm 0.116$) & 2.203($\pm 0.087$) & \textbf{1.924($\pm 0.015$)} & 1.060($\pm 0.013$) & 0.915($\pm 0.036$) & \textbf{0.761($\pm 0.011$)} \\ \midrule
\multirow{4}{*}{\textbf{D2}} & \textbf{1}  & 6.042($\pm 0.582$) & 3.905($\pm 0.163$) & \textbf{3.371($\pm 0.072$)} & 2.424($\pm 0.230$) & 1.926($\pm 0.114$) & \textbf{1.634($\pm 0.029$)} \\             & \textbf{4}  & 7.320($\pm 0.375$) & 4.663($\pm 0.132$) & \textbf{2.640($\pm 0.188$)} & 2.369($\pm 0.099$) & 1.807($\pm 0.024$) & \textbf{1.677($\pm 0.017$)} \\
             & \textbf{7}  & 8.131($\pm 0.512$) & 4.571($\pm 0.210$) & \textbf{3.168($\pm 0.164$)} & 2.493($\pm 0.138$) & 1.824($\pm 0.078$) & \textbf{1.615($\pm 0.092$)} \\
             & \textbf{10} & 6.572($\pm 0.219$) & 4.455($\pm 0.319$) & \textbf{2.508($\pm 0.096$)} & 2.324($\pm 0.151$) & 1.808($\pm 0.013$) & \textbf{1.707($\pm 0.010$)} \\ \bottomrule
\end{tabular}
\label{tab:mse}
\end{table*}

% \usepackage{booktabs}
\begin{table*}[t]
\centering
\caption{The CRPS (lower is better) results of DiffTCN and four baseline methods. (The reported results are the average performance of ten trials.)}
\begin{tabular}{cccccc}
\toprule
         & \multicolumn{1}{c}{\textbf{DeepAR}} & \multicolumn{1}{c}{\textbf{DeepFactor}} & \multicolumn{1}{c}{\textbf{Diffusion}} & \multicolumn{1}{c}{\textbf{TimeGrad}} & \multicolumn{1}{c}{\textbf{DiffTCN}} \\ \midrule
\textbf{D1}  & 0.065($\pm 0.007$)& 0.064($\pm 0.001$)  & 0.082($\pm 0.009$)  & 0.067($\pm 0.003$)   & \textbf{0.052($\pm 0.001$) }     \\
\textbf{D2}  & 0.096($\pm 0.008$)& 0.095($\pm 0.001$)  & 0.127($\pm 0.010$)  & 0.091($\pm 0.004$)   & \textbf{0.081($\pm 0.002$)}   \\ \bottomrule
\label{table:CRPS}
\end{tabular}
\end{table*}

% \begin{figure*} 
% \centering
% 	\subfloat[\label{fig:a}]{
% 		\includegraphics[scale=0.36]{20_20and20_21.png}}
% 	\subfloat[\label{fig:c}]{
% 		\includegraphics[scale=0.36]{18_20and18_21.png}}
% 	\subfloat[\label{fig:e}]{
% 		\includegraphics[scale=0.36]{test_set.png}}
% 	\\
% 	\subfloat[\label{fig:b}]{
% 		\includegraphics[scale=0.36]{16_20and16_22_2.png} }
% 	\subfloat[\label{fig:d}]{
% 		\includegraphics[scale=0.36]{19_20and19_22_2.png}}
% 	\subfloat[\label{fig:f}]{
% 		\includegraphics[scale=0.36]{test_set_2.png}}
% 	\caption{The histograms (distributions) of 100 inferred samples from DiffTCN (blue), TimeGrad (green) along with the ground truth (orange). Fig.~(a) and (b) are the results at two random time positions on dataset D1. Fig.~(c) is the aggregated distributions of all time positions on dataset D1. Fig.~(d) and (e) are the results at two random time positions on dataset D2. Fig.~(f) is the aggregated distributions of all time positions on dataset D2.}
% 	\label{fig:result} 
% \end{figure*}

We embarked on an insightful comparative analysis, juxtaposing our proposed Diffusion Temporal Convolutional Network (DiffTCN) against two analogous diffusion-based methodologies, Diffusion and TimeGrad. The empirical results, as delineated in Table~\ref{tab:mse}, present a comprehensive overview of the attained mean squared error (MSE) and mean absolute error (MAE) metrics. The methods exhibiting superior performance are denoted in bold typeface. Notably, DiffTCN consistently attains the highest precision across diverse prediction horizons, outperforming the baseline models that overlook crucial contextual cues. Of particular interest, TimeGrad melds the diffusion model with an encoder architecture founded upon a recurrent neural network (RNN), thereby accentuating the ability of Temporal Convolutional Networks. This advantage stems from the tangible augmentation of predictive precision facilitated by explicitly incorporating latent contextual dynamics, as we thoroughly analyzed in Section {\ref{sec:dynamics}}.

Presented in Table~\ref{table:CRPS}, a comprehensive exposition of the continuous ranked probability score (CRPS) reaffirms DiffTCN's preeminence among the four established baseline methods. This evaluation emphasizes the adeptness and appropriateness of our proposed approach for probabilistic time series prediction. Notably, we postulate that the notable reduction in CRPS is attributed to harnessing the remarkable capabilities inherent in the transformer layers, as introduced in Section~\ref{sec:archtecture}. These layers exhibit abilities to capture temporal intricacies and inter-dependencies, thereby fortifying the model's predictive capability concerning forthcoming QoS metrics.
It is worth highlighting that all results provided in the aforementioned tables are derived from aggregating outcomes from 10 distinct experimental runs. This noteworthy facet underscores the robustness and reliability of DiffTCN, as it consistently attains the outcomes displaying the least variability amidst diverse evaluations.

% In Figure~\ref{fig:result}, we provide the histograms (distribution plots) showcasing the inferred samples for Dataset D1 and D2. The visual depiction demonstrates the resemblance between the forecast distributions generated by DiffTCN and the actual data distribution, in contrast to the forecast distributions yielded by TimeGrad, the second-ranking method as per Table~\ref{table:CRPS}.
%
A discernible trend observed in the figures is the propensity of TimeGrad's forecast distribution to exhibit greater dispersion, indicative of DiffTCN's adeptness in mitigating prediction uncertainty by capturing their distributions more effectively.

\eat{
We began by contrasting our proposed approach, DiffTCN, with two similar diffusion-based methods, Diffusion and TimeGrad. The results of our experiments, shown in Table{\ref{tab:mse}}, outline the mean squared error and mean absolute error outcomes. The methods demonstrating superior performance are indicated in bold font. Evidently, DiffTCN consistently achieves the highest accuracy across all prediction horizons compared to the baseline models that disregard contextual information. In particular, TimeGrad melds the diffusion model with an encoder built upon a recurrent neural network (RNN). This observation indicates that the TCN element truly enhances the modeling of a more precise distribution. As we thoroughly analyzed in section {\ref{sec:dynamics}}, this advantage stems from the explicit fusion of latent contextual dynamics.
}
\eat{
Table~\ref{table:CRPS} showcases the outcomes for the continuous ranked probability score (CRPS). Among the four baseline methods, DiffTCN stands out with the most favorable CRPS results. These findings underline the efficacy and suitability of the proposed approach in the context of time series prediction.
Additionally, we hypothesize that the reduction in CRPS is attributable to leveraging the remarkable capabilities of the transformer layers introduced in section {\ref{sec:archtecture}}. These layers adeptly capture temporal correlations and dependencies, thereby bolstering the predictive ability for future QoS metrics.
}

%\section{Related work}
%\label{sec:related}
\input{related}

\section{Conclusion}
\label{sec:con}
%1. *recapitulate your finding*
%2. *list limitations*
%3. *implication for future research*
%time complexity
\eat{
This paper proposes a framework based on diffusion models for probabilistic QoS metrics forecasting in DTN, which overcomes the critical issues in existing deterministic regression methods. 
By incorporating the latent contextual dynamic, the proposed approach can capture data complexity to quantify the uncertainty of forecasts. Experimental results show that the proposed approach could perform better than popular time series forecasting methods. One promising future direction could be extending the approach to the out-of-distribution circumstance, where the inference is made on data with different distributions from the training data.}

This study has introduced a framework rooted in diffusion models for probabilistic QoS metrics forecasting within DTN, which effectively surmounts the pivotal challenges entrenched within prevalent deterministic regression techniques.
The assimilation of latent contextual dynamics within our framework engenders a heightened capacity to capture data complexity to quantify the uncertainty of forecasts. Empirical evaluations validate its efficacy in addressing real-world forecasting scenarios, which endorse the superiority of our approach vis-à-vis conventional time series forecasting methodologies.
One promising future direction could be extending our framework's capabilities to embrace the out-of-distribution scenarios, where the inference is made on data with different distributions from the training data. This prospect represents a natural evolution of our work, potentially yielding a more encompassing solution to complex forecasting challenges in dynamic and heterogeneous environments.

%\acknowledgements
%Acknowledgements.

\bibliographystyle{jaciiibibtex}
\bibliography{main, ref_legacy}

\end{document}

%% file: prob.tex
We formulate the QoS metrics prediction in DTN as a probabilistic multivariate time series forecasting problem, where one needs to forecast the distribution of future values up to a specific time horizon. The time horizon here is called \emph{forecasting horizon} or \emph{lead time}. 

Mathematically,
Let $\mathbf{X}=(\mathbf{x}_{1},\mathbf{x}_{2},...,\mathbf{x}_{t})\in\mathbb{R}^{m\times t}$ denote a multivariate time series, where $m$ is the number of variables in the time series and $t$ is the length of the time series. $\mathbf{x}_i=(x_{i,1},x_{i,2},\cdots,x_{i,m})$ is the measurements of the input multivariate time series at time $i$ which could be QoS metrics or other observable variables in this paper. 
To forecast the future distributions of potential values on the multivariate time series, one can sample the future values, denoted as $\mathbf{Y}=(\mathbf{y}_{t+1},\mathbf{y}_{t+2},...,\mathbf{y}_{t+p})\in\mathbb{R}^{n \times p}$, multiple times. Here $\mathbf{y}_{j}=(y_{j, 1},y_{j, 2},\cdots,y_{j, n})$ is the sampled measurements at time $j$, and $p$ is the forecasting horizon.

When $p$ is 1, it is evident that one can sample the subsequent value multiple times to obtain the corresponding distribution. When $p$ is greater than 1, one approach to accomplish this is explicitly training several prediction models for the steps smaller than $p$, which would considerably increase the computation cost. Alternatively, one may iteratively conduct a one-step-ahead prediction up to the specified time horizon, using the previously forecast values as the historical values for future predictions. The latter method does not require prior knowledge of the value of $p$ and applies to any arbitrary value of $p$ as long as the precision is adequate. This work employs the iterative approach to conduct one-step-ahead forecasting up to the chosen horizon.

%The value of the prediction horizon $p$ depends on the individual applications. If the actual value $p'$ of the prediction horizon in applications are different from the pre-defined

%and sometimes it not an easy task to decide its value beforehand. The solution is that we could use a pre-defined value of $p$ to train a prediction model. Let $p'$ denote the actual value of the prediction horizon. When $p' \leq p$, we can directly using the $p'$ values exiting in the prediction results for $p$. If $p' > p$, then we can adopt a similar strategy from the iterative method, which is to use previously predicted values as the context of the predicted values later. This strategy could be working for any arbitrary value of $p$, so long as the prediction accuracy is acceptable.

%By using predicted values instead of real observations, errors might be propagated and accumulated in the prediction model, which yields potential poor prediction performance.
%
%The other way is to train $p$ prediction models explicitly. Each of them only uses the context information $p$-step ahead, which considers much less context of the predicted values.

%We need to carefully select the appropriate prediction strategy, as well as the prediction horizon.
%

%Use an example to introduce the time series prediction

%And formulate the problem (including prediction horizon and how to handle random length of horizons.)

%% file: tcn.tex
\eat{
\begin{figure}[t]
    \centering
    \includegraphics[width=0.9\columnwidth]{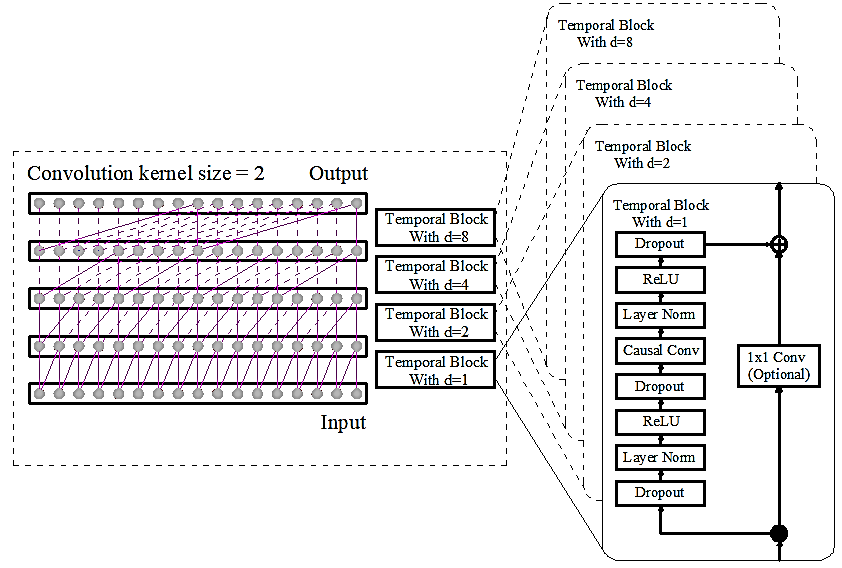}
    \caption{TCN Architecture}
    \label{fig:tcn}
\end{figure}
}
We propose to enhance the prediction capacity of the conditional diffusion model for QoS metric forecasting in DTN by incorporating state-of-the-art Temporal convolutional Networks (TCN) \cite{BaiTCN2018}. The primary reason behind our choice of TCN lies in its convolutional network's inherent parallelism and its seamless integration with the diffusion model. Despite the theoretical potential of recurrent architectures to capture an infinite historical context, TCNs stand out by exhibiting notably extended memory retention. This attribute makes them particularly well-suited for scenarios demanding a comprehensive historical perspective. This longer memory retention offered by TCNs is especially advantageous when combined with the diffusion model to capture intricate contextual dynamics. With TCNs' ability to retain extended historical context, they excel at capturing the nuanced patterns and dependencies in the data over extended periods. By leveraging the generalizability of the diffusion model and the tractability of the TCN, generative time series forecasting can be enhanced, leading to improved overall accuracy and robustness in estimating subsequent QoS metrics.

This paper captures the contextual dynamics in Section \ref{sec:cond} by TCNs. In sequence modeling, TCN is a popular method based on convolutional networks. As shown in the right side of Fig.~\ref{fig:all}, the major components in TCN include (1) causal convolution, which convolutes features across the time domain and prevents the information from leaking; (2) 1D fully-convolutional network (1D FCN), which maps the input to the output; (3) dilated convolution, which could extend the receptive field of causal convolution; and (4) residual temporal blocks, which deepen the overall network. 
These TCN components offer several benefits, such as customizable receptive fields, excellent parallelism, and a steady gradient descent during training.
The framework provided in this paper has no restriction on the neural model for latent contextual dynamics. Nevertheless, the main reason to select TCN is the parallelism of convolution networks and its efficient combination with the diffusion model. 

\eat{
indeed of the traditional recurrent neural model such as LSTM \cite{DBLP:conf/nips/HochreiterS96} and GRU \cite{DBLP:journals/corr/ChungGCB14} 
}

%% file: related.tex
\section{Related Work}
\label{sec:related}
\eat{
This section briefly introduces the research works in multivariate time series forecasting, covering traditional time series forecasting techniques and learning-based time series forecasting methods, followed by research efforts on QoS prediction in delay-tolerant networks.
If readers are interested in time series prediction using standard machine learning or deep learning approaches, they may reference several good textbooks and surveys\cite{forcasting1,brockwell2016introduction,2004.13408}.
}

\eat{
\subsection{Traditional methods for time series forecasting}

Forecasting time series has received substantial scholarly attention in a variety of applications\cite{dong_greedy_2015, DBLP:conf/dgo/NamP11}. Despite these efforts, the effectiveness of existing prediction algorithms needs to improve, with the difficulty of time series forecasting continuing to dominate and permeate the field. Thus, further research and development are necessary to address these ongoing challenges and improve forecasting capacities.

The statistical and stochastic process theories are the foundation of conventional time series prediction techniques. The Autoregressive Integrated Moving Average (ARIMA) \cite{doi:10.1080/01621459.1970.10481180} is a well-known prediction model that utilizes the inherent cyclical characteristics of time series data to provide accurate forecast predictions. Classical approaches struggle to produce solid results in the presence of multivariate time series exhibiting stochastic fluctuations and uncertainties, especially when confronted with complex, nonlinear multivariate time series.

Other prominent techniques include Support vector regression (SVR)\cite{DBLP:journals/sac/SmolaS04}, Gaussian Process (GP)\cite{Chu2004Gaussian}, and more recently, Prophet\cite{taylor2018forecasting}. SVR constructs a linear regressor by minimizing a specified cost function and projects the input onto a higher-dimensional feature space. GP is a non-parametric approach for predicting time series. Prophet is a modular regression model whose parameters may be adjusted and understood in light of existing domain information.

\eat{
Support vector regression (SVR) represents another prominent technique in the realm of time series forecasting\cite{DBLP:journals/sac/SmolaS04}. SVR establishes a linear regressor by minimizing a designated cost function and projects the input into a higher-dimensional feature space. This approach allows linear properties in the high-dimensional space to correspond to the nonlinear time series in the lower-dimensional feature space. The meticulous selection of hyperparameters, encompassing the kernels for basis functions, is essential for SVR. Moreover, substantial storage capacity is required to address the convex optimization problem inherent in high-dimensional spaces. Alternative non-parametric time series prediction methodologies, such as the exemplary Gaussian Process (GP), can be employed to predict continuous function values, further expanding the repertoire of forecasting techniques available to researchers and practitioners\cite{Chu2004Gaussian}.

Prophet, a modular regression model recently introduced by Facebook, employs adjustable parameters that can be fine-tuned and interpreted in light of prior domain knowledge. This model demonstrates efficacy for time series data exhibiting pronounced periodic characteristics. However, in instances where the data is characterized by substantial randomness and volatility, the prediction accuracy of Prophet cannot be unequivocally assured. 
}

\subsection{Learning-based approaches for time series forecasting}

Nonlinear multivariate time series frequently exceed the capability of conventional time series prediction algorithms due to their complexity. Through the implementation of deep learning techniques, a variety of applications, including speech recognition, pattern identification, natural language translation, and picture recognition, have witnessed significant advances. Due to the successful development of sophisticated deep learning architectures \cite{ijcai2020-693} and the increased parallel processing made possible by high-performance GPUs  \cite{DBLP:journals/candie/ChengG19}, time series forecasting has also made significant strides \cite{WangHW12,abs-1808-09794}.

Recurrent architectures are helpful as a starting point for tasks such as sequence modeling. However, traditional RNNs confront difficulties associated with disappearing and bursting gradients during training \cite{DBLP:journals/corr/abs-1211-5063}. Long Short-Term Memory (LSTM) networks \cite{DBLP:conf/nips/HochreiterS96}, which include a selective memory function, successfully solve gradient-related concerns and demonstrate their sequence modeling potential. Providing adequate optimization processes while training RNNs with LSTM cells is crucial \cite{pmlr-v28-pascanu13}. Residual Networks (ResNet) \cite{DBLP:conf/cvpr/HeZRS16} inspire RNN designs, and adding residual connections into RNNs provides several benefits, such as reduced number of hidden units, quicker training, and enhanced prediction accuracy. The Gated Recurrent Unit (GRU) \cite{DBLP:journals/corr/ChungGCB14}, a more streamlined variation of LSTM, demonstrates performance equivalent to that of LSTM.

Temporal Convolutional Networks (TCN) \cite{DBLP:journals/corr/abs-1803-01271} exhibit exceptional prediction skills while limiting the drawbacks of RNNs. Deep learning-based algorithms are effective at extracting latent characteristics from multivariate time series, taking into account both long-term and short-term dependencies, resulting in superior prediction performance compared to conventional time series prediction techniques. This progress in deep learning approaches highlights the potential for the continued growth of predictive analytics across a variety of application areas.

DeepAR \cite{SALINAS20201181} can produce accurate probabilistic forecasts for time series.
As the proposed model in this paper, DeepAR is an autoregression model, except that it is based on a recurrent neural network. The authors demonstrate that deepAR can handle widely-varying scales through rescaling and velocity-based sampling, generate calibrated probabilistic forecasts with high accuracy and learn complex patterns such as seasonality and uncertainty that grew over time from the data. 
DeepFactor \cite{pmlr-v97-wang19k} is a hybrid model that combines the benefits of classical time series models and deep neural networks to produce probabilistic forecasts for large collections of time series. The authors develop an efficient and scalable inference algorithm for non-Gaussian likelihoods that generally applies to normally distributed probabilistic models, such as state space models and Gaussian Processes. 
TimeGrad \cite{rasul_autoregressive_2021} is an autoregressive model for multivariate probabilistic time series forecasting. TimeGrad estimates the gradient of the data distribution at each time step using diffusion models. However, unlike the one in this paper, TimeGrad adopts a recurrent neural network whose hidden representations serve as the input of diffusion models.

}

\addtolength{\topmargin}{0.03in}
%\subsection{QoS metric forecasting in Delay-tolerant networks}
This section will briefly introduce the existing research on QoS metric prediction in delay-tolerant networks. Readers interested in time series prediction using deep learning approaches may refer to related textbooks and surveys \cite{forcasting1,2004.13408}.
%brockwell2016introduction
There are some ad hoc research efforts about QoS metric forecasting, which can improve the network performance in terms of latency, throughput, energy consumption, and dependability.%\cite{pundir_quality--service_2021}. 

Abdellah et al.  \cite{abdellah_machine_2021} studied the problem of network traffic prediction by combining AI with information networks. They used a nonlinear autoregressive with external input (NARX)-enabled recurrent neural network (RNN) to predict internet delays.
Nagai et al. \cite{nagai_wireless_2022} constructed a wireless sensor network testbed to study network delay outdoors while considering the Line-of-Sight (LoS) scenario. They used Long Short-Term Memory (LSTM) to forecast the delays.
Ghandi et al. \cite{ghandi_non-negative_2022} formulate the network delay prediction problem as a nonnegative matrix factorization problem with piece-wise constant coefficients of the approximate instantaneous data representation. 

The continual data transmission of energy-constrained sensor nodes hampers the longevity and performance of wireless sensor networks. Engmann et al. \cite{s_yellampalli_applications_2021} investigated various approaches to data prediction in wireless sensor networks, including stochastic approaches, time series forecasting, and machine learning approaches. 
Greidanus et al. \cite{roig_greidanus_novel_2021} presented two innovative control techniques that permit localized delay compensation for grid-connected and independent inverters. The core is a prediction policy based on a robust delay mitigation range.
Tian and Wang \cite{balsamo_prediction_2021} presented a predictive control compensation strategy for networked control systems to deal with random time delays. They employed several delay compensation mechanisms to enhance the control impact based on the delays between the actual control signal and the controller output at the historical sampling time.

Accurately predicting network conditions and making real-time adjustments can improve network communication quality. Ma et al. \cite{ma_research_2021} employed the ARMA prediction model to anticipate the network data and the optimized BP neural network to predict the network activities of an intelligent production line.
Wang et al. \cite{wang_soft_2021} applied a soft sensor network with dynamic time-delay estimation in a wastewater treatment plant. They proposed a weighted relevance vector machine model based on dynamic time-delay estimation to implement quality variable prediction.

\eat{
There are some ad hoc research works about QoS metric forecasting.
Due to mobile sensor nodes' small and dynamic nature, it is difficult to satisfy the requirement for Quality-of-Service (QoS) in a real-world setting. \cite{pundir_quality--service_2021} examines the QoS metric based on WSN's layered architecture. In addition, the proposed statistical study shows that the QoS metric improves the network's performance in terms of latency, throughput, energy consumption, and dependability. Finally, machine learning techniques are examined in light of predicting QoS indicators to improve the overall network performance.

Nagai et al. \cite{nagai_wireless_2022} constructed a wireless sensor network testbed In order to study wireless sensor network delay in an outdoor setting while taking Line-of-Sight (LoS) scenario into consideration. They used Long Short-Term Memory to forecast the delay by long short-term memory (LSTM).

Abdellah et al. \cite{abdellah_machine_2021} study the problem of network traffic prediction by combing AI with information networks, which is an essential strategy for ensuring security, dependability, and quality of service (QoS) requirements. They use a nonlinear autoregressive with external input (NARX)-enabled recurrent neural network (RNN) to conduct time series prediction for IoT and tactile internet delays.

%It is now applicable to a number of different tasks, including network monitoring, resource management, traffic control, bandwidth allotment, network intrusion detection, etc. 

Ghandi et al. \cite{ghandi_non-negative_2022} formulate the network delay prediction problem as a nonnegative matrix factorization problem with piece-wise constant coefficients of the approximate instantaneous data representation. The method takes advantage of the substantial geographical and temporal correlation present in network delay data and shows effectiveness in the experimental results on two data sets.

The longevity and performance of WSNs are hampered by the continual data transmission of energy-constrained sensor nodes. Engmann et al. \cite{s_yellampalli_applications_2021} investigated various approaches of data prediction in wireless sensor networks, including stochastic approaches, time series forecasting, algorithmic approaches, and machine learning approaches. Data prediction could help to schedule nodes' sleep periods for lifetime improvement. 

Improving network communication quality depends on the ability to accurately predict network conditions and make real-time adjustments. Ma et al. \cite{ma_research_2021} provide an algorithm for predicting the network condition of an intelligent production line that employs the ARMA prediction model to anticipate the network data and the optimized BP neural network to calculate and predict the full network activity.

Wang et al. \cite{wang_soft_2021} models a soft sensor network with dynamic time-delay estimation and applied the network in the wastewater treatment plants. Because of the variable transmission times of signals in actual industrial processes, it is no longer possible to reliably extract the characteristics of dynamic time delays using the conventional static time-delay estimation methods. In order to implement dynamic characteristic extraction and quality variable prediction, a weighted relevance vector machine model based on dynamic time-delay estimation is proposed in this study.

Greidanus et al. \cite{roig_greidanus_novel_2021} present two innovative control techniques that permit localized delay compensation for grid-connected and independent inverters. The key of the suggested techniques is a robust delay mitigation range by a prediction policy, where the prediction horizon of conventional controllers is rather narrow in comparison to massive communication delays and information unavailability.

Tian and Wang \cite{balsamo_prediction_2021} present a predictive control compensation strategy for networked control systems to deal with random time delay. This work employs several input and output time-delay compensation mechanisms to enhance the control impact. Combining the historical output value and the control variable, the fast implicit generalized predictive control algorithm is utilized to create the predictive controller to adjust the input channel time delay. The output control variable of the predictive controller is adjusted for the output channel time delay by anticipating the error between the actual control signal and the controller output at the historical sampling time.
}